\documentclass[journal,transmag]{IEEEtran}
\usepackage{subcaption}
\usepackage{graphicx}

\usepackage{layouts}

\usepackage{cite}

\ifCLASSINFOpdf
\else
\fi
\usepackage[hidelinks]{hyperref}
\makeatletter
\renewcommand{\section}{\@startsection {section}{1}{\z@}%
             {-3.5ex \@plus -1ex \@minus -.2ex}%
             {2.3ex \@plus .2ex}%
             {\normalfont\large\scshape\bfseries}}

\renewcommand{\subsection}{\@startsection{subsection}{2}{\z@}%
             {-3.25ex\@plus -1ex \@minus -.2ex}%
             {1.5ex \@plus .2ex}%
             {\normalfont\normalsize\scshape\bfseries}}
\makeatother
\begin{document}
%
\title{Process Knowledge-Infused AI: Towards User-level Explainability, Interpretability, and Safety}


\author{\IEEEauthorblockN{Amit Sheth,~
 Manas Gaur,~
Kaushik Roy,~
Revathy Venkataraman,~
Vedant Khandelwal}
\IEEEauthorblockA{AI Institute, University of South Carolina, Columbia, SC, USA, 29201}
}

\maketitle

\IEEEdisplaynontitleabstractindextext

%
\IEEEpeerreviewmaketitle

\begin{figure*}[!htbp]
    \centering
    \includegraphics[width=0.7\textwidth]{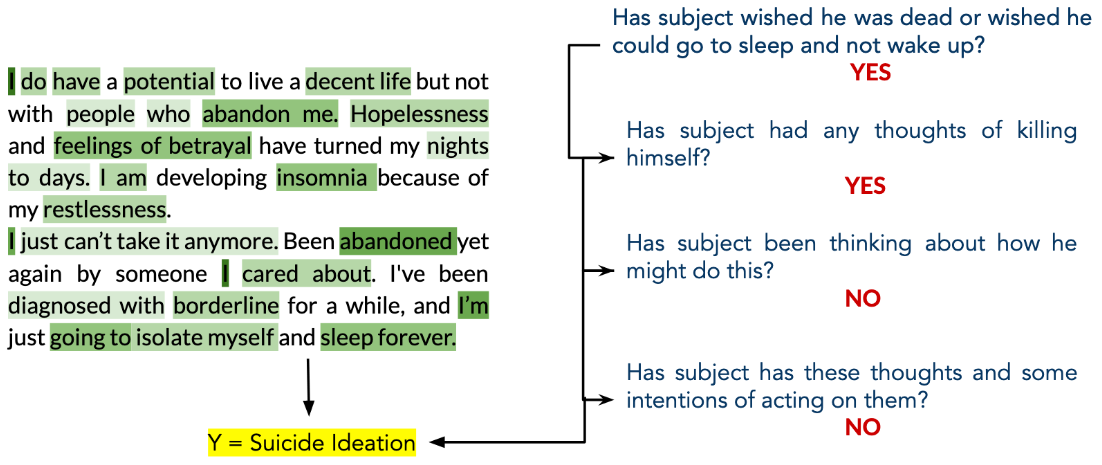}
    \caption{An illustration of a classification task that benefits from process knowledge. Here, an AI model using a process knowledge structure would consume the user's input, extract conceptual cues that can answer questions in process knowledge, and provide a classification label. The benefit of the process knowledge infusion in AI is the deterministic nature that it enforces in AI to achieve user-level explanations, handle uncertainty, and be safe. The figure illustrates this process in assessing suicide risk severity using a partial sequence of questions from the Columbia Suicide Severity Rating Scale (C-SSRS) (\url{https://tinyurl.com/Posner-CSSRS}). The highlighted text on the left is concept phrases that contribute to the yes/no in the C-SSRS questions. The graph on the right is Process knowledge in C-SSRS.}
    \label{fig:text-to-questions}
\end{figure*}

\IEEEPARstart{A}{I} systems have been widely adopted across various domains in the real world. However, in high-value, sensitive, or safety-critical applications such as self-management for personalized health or food recommendation with a specific purpose (e.g., allergy-aware recipe recommendations), their adoption is unlikely. Firstly, the AI system needs to follow guidelines or well-defined processes set by experts; the data alone will not be adequate. For example, to diagnose the severity of depression, mental healthcare providers use Patient Health Questionnaire (PHQ-9). So if an AI system were to be used for diagnosis, the medical guideline implied by the PHQ-9 needs to be used. Likewise, a nutritionist's knowledge and steps would need to be used for an AI system that guides a diabetic patient in developing a food plan. Second, the BlackBox nature typical of many current AI systems will not work; the user of an AI system will need to be able to give user-understandable explanations, explanations constructed using concepts that humans can understand and are familiar with. This is the key to eliciting confidence and trust in the AI system. For such applications, in addition to data and domain knowledge, the AI systems need to have access to and use the Process Knowledge, an ordered set of steps that the AI system needs to use or adhere to. The use of process knowledge also aids in developing an AI system that has other important features for such demanding applications, such as safety, which is critical to ensure the AI system remains within the bounds and does not cause harm to a user - something that current AI systems using Large Language Models have failed to do. This article will use two demanding applications – mental health triaging and cooking recipes for diabetes to show what process knowledge they need and how process knowledge can be modeled and then infused into AI algorithms to achieve the outlined features. We also discuss user-level explainability, how to support safety and present metrics for performance evaluation.

\section*{Benefits and importance of process knowledge}
\noindent Benchmarking datasets that assess the natural language understanding capabilities of large language models fall short in accelerating models to achieve user-level explainability, safety, uncertainty, and risk handling (\url{https://tinyurl.com/KiL-MentalHealth-NLU}) \cite{sheth2021knowledge}. These challenges are associated with the limitations of AI in restricting its learning tasks to classification and generation, which are single shots. In comparison, real-world applications demand an orchestrated response going through a multi-step process of learning the high-level needs of the user, then drilling down to specific needs, and subsequently yielding a structured response having a conceptual flow. For example, triaging patients in mental health requires clinical process knowledge manifested in a clinical questionnaire. Figure \ref{fig:text-to-questions} illustrates a scenario where the agent maps user input to a sequence of yes or no questions to compile suicide risk severity. The agent can keep track of user-provided cues and ask appropriate follow-up questions through these ordered sets of questions. Upon receiving the required information to derive appropriate severity labels, the agent’ outcome can be explained to mental healthcare providers for appropriate intervention. Similar but more complex applications include using ADOS (Autism Diagnostic Observation Schedule) to evaluate children with autism or using MoCA (Montreal Cognitive Assessment) score to measure the cognitive decline in post-stroke Aphasia patients \cite{newman2021aging}
. To train conversational agents for such functionality requires specialized datasets grounded in the knowledge that enables AI systems to exploit the duality of data and knowledge for human-like decision-making \cite{sheth2019shades} (\url{https://tinyurl.com/duality-data-knowledge}). Further, to develop agents that learn from such process knowledge-integrated datasets we require interpretable and explainable learning mechanisms (\url{https://tinyurl.com/petrinet-workflow}). These learning mechanisms have been characterized under the umbrella of Knowledge-infused Learning.

\noindent\rule[0.5ex]{\linewidth}{1pt}
  \textit{Knowledge-infused Learning (KiL) is a class of Neuro-Symbolic AI techniques  is that utilize a variety of knowledge (lexical, linguistic, domain-specific, common-sense, process knowledge, and constraint-based) in different forms and abstractions into deep neural networks. It improves upon data-centric statistical learning to reduce training, reduce computing needs, and broaden coverage, resulting in improved performance, safety and model interpretation, and providing user-level explanations.}
\noindent\rule[0.5ex]{\linewidth}{1pt}

\section*{Shades of Process Knowledge Infused Learning}
\begin{figure}[!htbp]
     \centering
     \begin{subfigure}[b]{0.3\textwidth}
         \centering
         \includegraphics[width=\textwidth]{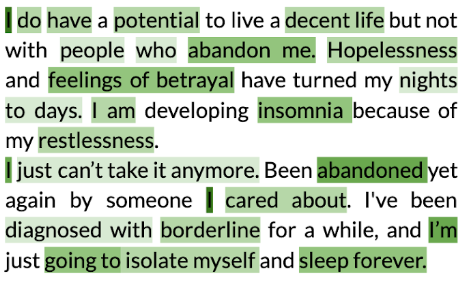}
         \caption{}
         \label{fig:y equals x}
     \end{subfigure}
     \hfill
     \begin{subfigure}[b]{0.49\textwidth}
         \centering
         \includegraphics[width=\textwidth]{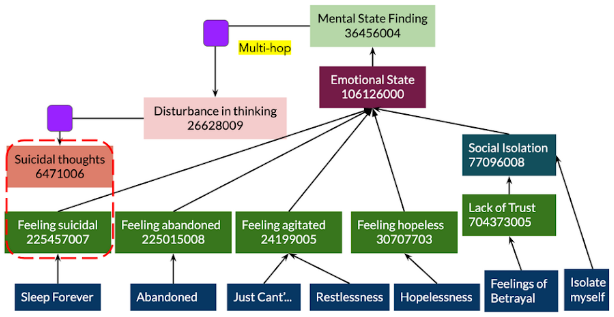}
         \caption{}
         \label{fig:three sin x}
     \end{subfigure}
        \caption{An illustration of user-level explainability using the important conceptual phrases identified by a deep learning model trained using the method in (\url{https://tinyurl.com/contextual-classify-reddit}). Highlighted phrases in (A) are queried in SNOMED-CT, thus forming a contextual tree. The formation of this tree is stopped when a node is hit that has high similarity to either leaf nodes or one-hop parent nodes. The resulting tree is shown in (B). The numbers in the boxes are SNOMED-CT IDs.}
        \label{fig:user-level-ex}
\end{figure}

\noindent KiL aligns with the third phase of DARPA to promote contextual adaptation in AI systems for user-level explanations. An AI system trained with knowledge infusion techniques provides forms of explanations by querying, traversing, and mapping the high importance features to concepts in a knowledge graphs. Figure \ref{fig:user-level-ex} illustrates the user-level explanations provided by an AI system infused with knowledge that highlights concept phrases in the input text. These concept phrases are used traverse a knowledge graph, which in figure \ref{fig:user-level-ex} is SNOMED-CT. Along with Figure \ref{fig:text-to-questions} that explains process knowledge infusion contributing to reasonable path towards classification, Figure \ref{fig:user-level-ex} provides additional user-level explanation. Within the three forms of knowledge-infusion under KiL (i.e., shallow, semi-deep, and deep \cite{sheth2019shades}), process knowledge infusion develops a new and complementary set of methods, datasets, and evaluation methods under semi-deep and deep knowledge infusion. 

Furthermore, any AI systems trained with such methods and over such datasets can also handle uncertainty and risk. They can establish the connection between the input and output, answering, "why such an outcome, given an input?". The AI systems are context-sensitive rather than opinionated based on only the input data, i.e. a partial representation of the world. The structure and order provided by using process knowledge allows the end-users a control over the AI system. Moreover, in-process knowledge in a particular domain and for a particular task (classification or generation), an AI system with a method that makes the model adaptable to process knowledge can make the system transferable across tasks. The subsequent sections will provide a concrete definition of process knowledge and its use in understanding and controlling AI models. With a focus on natural language generation (NLG), we will conceptually describe methods for infusing process knowledge into statistical AI systems. Thereafter, we provide use-cases in the domain of mental health (continuing with Figure \ref{fig:text-to-questions} and Figure \ref{fig:user-level-ex}) and cooking.  

\section*{Process Knowledge and Its Infusion into Statistical AI }
\noindent Process knowledge is an ordered set of information that maps to evidence-based guidelines or categories of conceptual understanding to experts in a domain. For instance, The American Academy of Family Physicians (AAFP) develops clinical practice guidelines (CPGs) that serve as a framework for clinical decisions and supporting best practices. CPG allows systematic assessment to optimize patient care. On the other hand, U.S. Departments of Agriculture (USDA) and Health and Human Services (HHS) develops Dietary Guidelines for Americans (\url{https://tinyurl.com/american-dietary}) that serves as an recommendation for meeting nutrient needs, promote health, and prevent disease. An AI system adapted to process knowledge can handle uncertainty in prediction, and the predicted outcomes are safe and user-level explainable. Further, an AI system can consider process knowledge as meta-information to capture the sequential context necessary for carrying out a structured conversation. Also, it allows the developer of the AI system to probe the internal decision-making of AI systems using application-specific guidelines or specifications that inform the synchrony between the end-users thought process and the model’s functioning.

This unique form of knowledge differs from other forms of knowledge in the following manner: (a) knowledge graph: it is structured but not ordered. Knowledge graphs can support context capture but cannot enforce conceptual flow (\url(https://tinyurl.com/KI-summarization)). (b) Semantic lexicons: this is a flattened form of knowledge graph that makes deep language models context-sensitive and add constraints but cannot enforce conceptual flow \cite{libben2021lexicon}
. (c) Ontologies are curated schematic forms of knowledge graphs with classes, instances, and constraints. Thus, ontologies can provide stricter control over context and constraints. If defined, an ontology can enforce order in question generation using deep language models \cite{stasaski2017multiple}
. Process knowledge is represented differently for different applications. For instance, to assess the severity of suicide risk, the process knowledge used is C-SSRS, which is similar to a flow chart. On the other hand, the GAD-7-based process knowledge is used to assess anxiety severity which has a flattened structure (see Figure \ref{fig:forms-of-process-knowledge}). DASH Diet based process knowledge can be used to assess the dietary intake of hypertension patients and also recommend meals.   These characteristic properties of process knowledge and its infusion into statistical AI would yield a new class of neuro-symbolic algorithms that would drive the question: 

\begin{figure}
     \centering
     \begin{subfigure}[b]{0.5\textwidth}
         \centering
         \includegraphics[width=\textwidth]{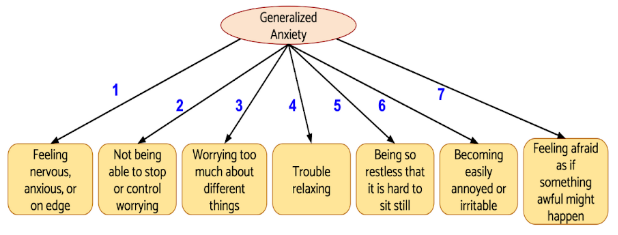}
         \caption{}
         \label{fig:y equals x}
     \end{subfigure}
     \hfill
     \begin{subfigure}[b]{0.3\textwidth}
         \centering
         \includegraphics[width=\textwidth]{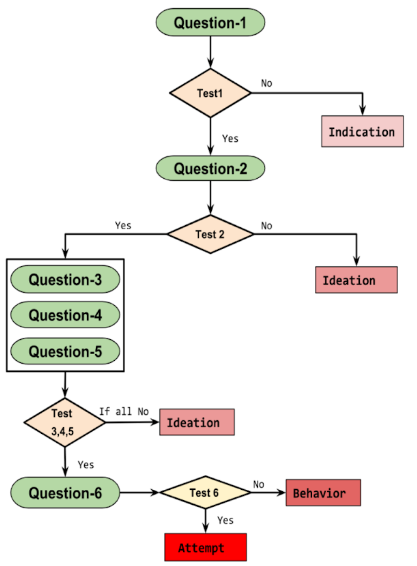}
         \caption{}
         \label{fig:three sin x}
     \end{subfigure}
        \caption{Illustration of process knowledge for different purposes.}
        \label{fig:forms-of-process-knowledge}
\end{figure}

\noindent\rule[0.5ex]{\linewidth}{1pt}
\textit{  What if we could use the annotator's labels and the process or guidelines used to label them and explicitly control the learning of a model to recover the guideline or process (instead of implicitly).}
\noindent\rule[0.5ex]{\linewidth}{1pt}

Such an algorithm would, by design, be explainable and emulate the human model of similarity between data points. For the task of classification, a process knowledge-infused AI system would solicit the use of interpretable machine learning algorithms (e.g., Decision Trees, Random Forest) that can enforce structure in decision making over traditional deep language model-based classification (\url{https://tinyurl.com/PK-iL-suicide}).

\begin{figure*}[!htbp]
\centering
   \begin{subfigure}{0.6\linewidth}
   \centering
   \includegraphics[width=\linewidth]{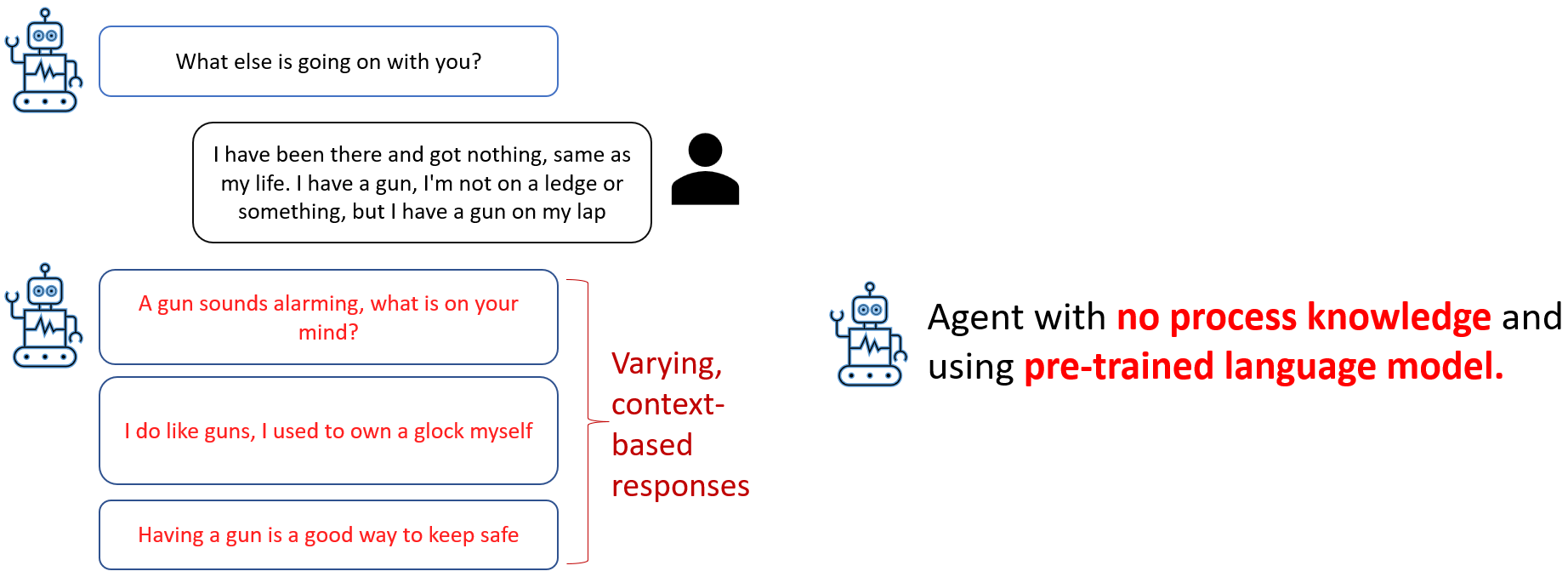}
   \caption{Extreme Behavior Detection}
   \label{fig:dragratio} 
\end{subfigure}
\hfill
\begin{subfigure}{0.25\linewidth}
   \centering
   \includegraphics[width=\linewidth]{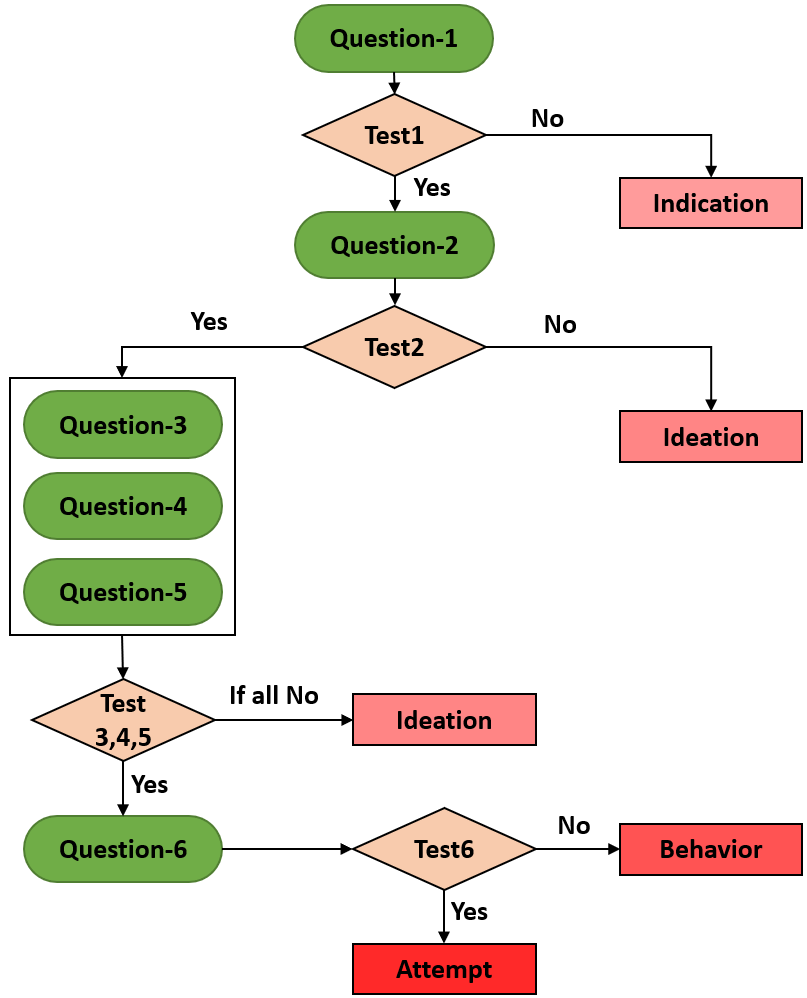}
   \caption{Process Knowledge to assess suicide risk severity}
   \label{fig:dragratio2}
\end{subfigure}
\\[\baselineskip]
\begin{subfigure}[H]{0.65\linewidth}
   \centering
   \includegraphics[width=\linewidth]{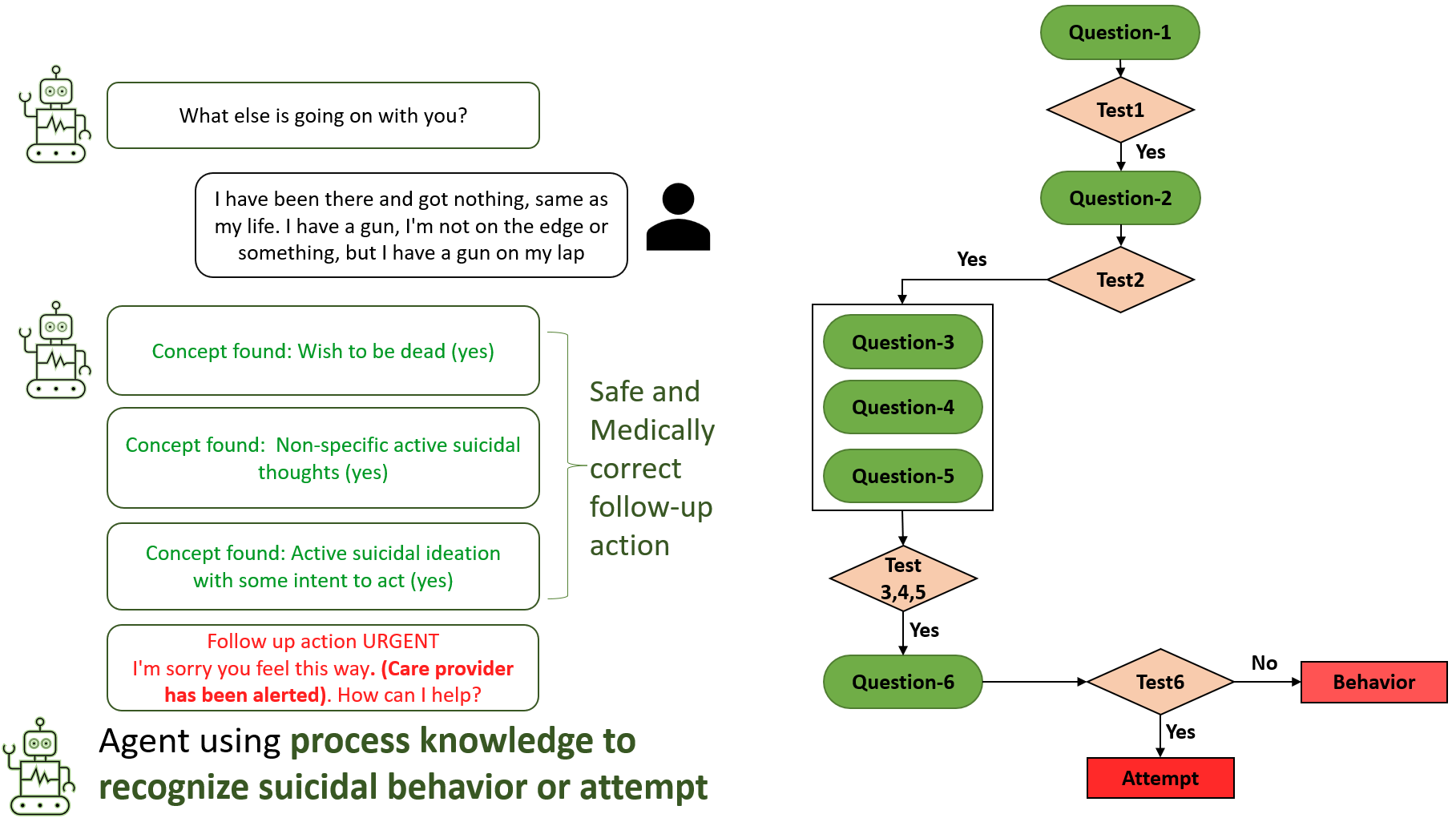}
   \caption{Model utilize process knowledge for safe and medically correct follow up generation question}
   \label{fig:dragratio3}
\end{subfigure}
\centering
\caption{An illustration of safety in conversational artificial intelligence. It explains why process knowledge is needed to avoid unsafe conversations and make models interpretable and explainable.}
\label{fig:safety}
\end{figure*}

In NLG, the biggest concern with deep generative language models is that they hallucinate when either asking questions or providing responses in a conversational setting. Along with the issue of hallucination, there have been extensive study about the inappropriate and unsafe risk behaviors of language models (\url{https://tinyurl.com/LaMDA-dialog}). Efforts to pair these language models with passage retrievers and rankers have been proposed to control incoherent, irrelevant, and factually incorrect responses and questions;  however, the order, like the one defined in process knowledge, is far from being realized \cite{glassre2g}
. Such process knowledge-based NLG is even more crucial in the field of healthcare NLP, where each response from the agent can have severe consequences. These concerns are further discussed with the help of two use cases: Mental Health and Food Domain. 

\section*{Mental Health Use-Case}

\noindent AI has contributed to the domains of drug research, customized medicine, and patient care monitoring and has the potential to aid physicians in making better diagnoses. However, when AI is used in health care, various dangers and problems might arise at the individual, macro, and technological levels (e.g., awareness, education, trust), as well as at the macro-level (e.g., legislation and rules, risk of accidents due to AI faults) (e.g., usability, performance, data privacy, and security). In the context of mental healthcare, conversational agents are prone to unsafe generations that can harm the user or engage in a conversation involving escalation in the severity of medical conditions. 

Figure \ref{fig:safety} illustrates a pipeline wherein (A) the deep statistical language model pre-trained on open domain corpus when tasked to converse with a user in a mental healthcare setting generates questions that it sees online. (B) Such questions are not what a mental healthcare provider would ask. If we utilize a clinical guideline, in this case, C-SSRS, the model can measure the safety of the generated question before asking. (C) Figure 4 shows a process over the detailed process knowledge that an AI agent followed to control its question generation and ask medically correct questions. A recent study from Roy et al. details this approach using C-SSRS, and Gupta et al. detail this approach using GAD-7 and PHQ-9, which are clinical guidelines to check whether the user is a patient of an anxiety disorder (GAD-7) or clinical depression (PHQ-9) \cite{roy2022process, gupta_2022_learning}.

\subsection*{Process Knowledge as Constraints}

\begin{figure*}[!htbp]
    \centering
    \includegraphics[width=0.75\textwidth]{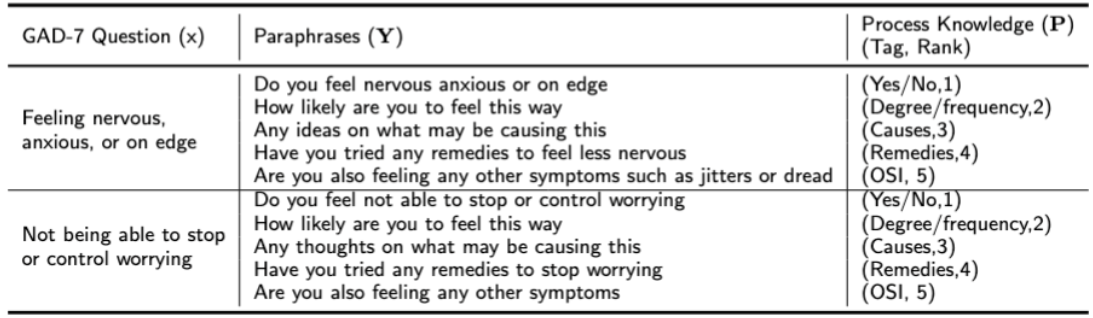}
    \caption{This is an example of how a process knowledge-integrated dataset is constructed in collaboration with mental healthcare providers. The leftmost column presents example questions mental healthcare providers (MHPs) asked. The MHPs provided Tag and Rank shown in the rightmost columns representing process knowledge. The middle column provides a series of questions gathered using Google SERP API (\url{https://tinyurl.com/G-SERP-api}) and Bing Search API (\url{https://tinyurl.com/bing-search-api}) logically ordered by MHPs.}
    \label{fig:process-knowledge-rules}
\end{figure*}

\noindent Some more ways in which process knowledge can be infused to add constraints and improve NLG of the current AI methods are:. 
\begin{itemize}
    \item Textual Entailment Constraints (TEC) is a directional relationship between sentences in a response or questions. If the two sentences share semantic relations and logically agree, they are entailed. If the two sentences are synonymous based on the entities they contain, they are neutral. If the second sentence refutes the information in the first sentence, they are contradictory. Such constraints are manifestations of process knowledge in clinical practice. In machine-understandable form, we can model them as Rules containing Tags and Rank (see Figure \ref{fig:process-knowledge-rules}). 
    \item Rules (Tag and Rank): These rules can help structure the question generation process, which is random and unsafe in current state-of-the-art NLG models (\url{https://tinyurl.com/adaptive-education}). For instance, if the conditional probability function within an AI model, defined as $P(\hat{Q}_{k+1}|\hat{Q}_k)$ is augmented with a Tag containing the following labels: \{Yes/No, Degree/Frequency, Causes, Treatment/Remedies\} then the model can learn to follow a definite process:
    \begin{itemize}
        \item If $\hat{Q}_k$ is Yes then $\hat{Q}_{k+1}$ is about Degree/Frequency
        \item If $\hat{Q}_k$ is Degree/Frequency then $\hat{Q}_{k+1}$ is about Causes
        \item If $\hat{Q}_k$ is Causes then $\hat{Q}_{k+1}$ is about Treatment/Remedies
        \item If $\hat{Q}_k$ is Treatment/Remedies then $\hat{Q}_{k+1}$ ask about Information on Other Side Effects
    \end{itemize}
    Here, $\hat{Q}_{k+1}$  is the next generated question given $\hat{Q}_k$, a previous generated and accepted question.
Further utility of constraints-based process knowledge infusion in AI is detailed in another application involving food recipe recommendation, wherein the constraints are defined based on allergens. 

\end{itemize}

\section*{Food Domain Use-Case}
\noindent A conversational system to manage diet can help patients in various applications such as hypertension and diabetes \cite{yagcioglu2018recipeqa}
. In most of the scenarios the interactions between user and system involve factual queries (e.g., Can you order a falafel for me? What are the sides offered with falafel? etc.). However, the challenge lies in how recommendations can be adapted to user preferences and context. It is still an open question \cite{pecune2022designing}. Further, how can recommendations be provided when users do not ask factual questions (e.g., Can you suggest some food that helps me control my calorie intake? I want to lose weight. What should I eat for lunch?). In case of Hypertension, patients need a nudge to switch towards healthy food habits. A nutrition management system can aid and assist them in this process. In such a scenario, when a user asks the following question to an agent: “Can you recommend dishes that are calorie efficient?” if the agent is augmented with the internet, it would accurately respond to the following related questions (or people-also-ask questions) : (a) “Are restaurants required to put calories on menus,” (b) “Are calorie recommendations accurate,” (c) “Should I eat less than my recommended calories?”, and (d) “What food can you recommend?”. Moreover, top-2 searches on google for the user query are (a) Cut lots of calories  and (b) How to lose weight eating more food, which are not relevant to the user query. There are two fundamental problems here: (a) The AI system behind these recommendations is confused about whether “calorie efficiency” is positive or negative. (b) The AI system fails to bridge the gap between dishes and calorie efficiency. Furthermore, a response to such a question is dependent on the time of the day: breakfast, lunch, or dinner. A process knowledge-based conversational agent would generate the following information-seeking questions: (a) Do you have any preference in cuisine? (b) Do you want to know about low-calorie food in this cuisine for breakfast/lunch/dinner? (c)  Do you want me to book reservations for restaurants that have this cuisine? (d) Do you want me to save your preferences? If the answer to (b) question is no, then an alternate path in process knowledge is triggered. Here, process knowledge is the procedure for recommending and ordering food. Moreover, the agent can benefit from the 2015-2020 Dietary Guidelines for Americans to emphasize overall healthy eating patterns supported by five food groups: fruits, vegetables, grains, protein foods, and dairy (\url{https://tinyurl.com/Dietary-Guidelines}).

\begin{figure*}[!ht]
    \centering
    \includegraphics[width=0.9\textwidth]{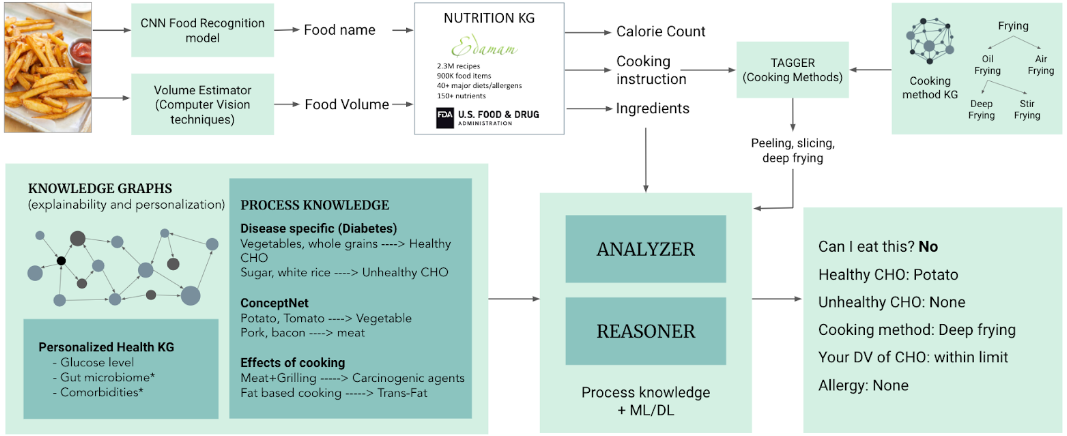}
    \caption{An illustration of the infusion of process knowledge in food recommendation agents.}
    \label{fig:food-recommendation}
\end{figure*}

Similarly, for type-I Diabetes, patients need to monitor carbohydrate (CHO) intake for their insulin dosage, hence the source of CHO determines whether a given food item is advisable. CHO count due to the fibers present in vegetables and fruits are considered healthy, whereas CHO from added sugars, white rice, and pasta are considered unhealthy (\url{https://tinyurl.com/Mayo-Diabetes-plan}). Existing models advise meals based on the daily value of the CHO limit of an individual. In this case, the CHO count of a recipe derived from added sugar will be recommended by the agent if it is within the daily CHO limit of an individual. This can have severe effects on an individual's health. By infusing the process knowledge of diabetic dietary guidelines into the learning process, the agent can learn to advise appropriate meals and generate explanations to enhance interpretability and safety (see Figure \ref{fig:food-recommendation}).

Along with above two scenarios, the nutrients content of the food change based on the adverse effects of the cooking actions on the final cooked food item such as nutrition loss or the introduction of harmful elements. To add, the dietary restrictions for each chronic disease have respective guidelines. Hence in this scenario, two kinds of process knowledge, adverse effects of cooking actions combined with ingredients and dietary guidelines for chronic conditions, are involved in generating explanations, improving interpretability and safety of food recommendation agents.

\subsection*{Process Knowledge as Constraints} 
\noindent In addition to specific dietary guidelines for chronic conditions, cooking actions produce adverse effects. For example, the ingredients for potato fries involve potatoes, oil, salt, pepper, and other seasonings. These are advisable ingredients as per dietary guidelines for diabetes. However, the cooking action is deep-frying which produces trans-unsaturated fatty acids. The trans-unsaturated fatty acids are not advisable for any chronic diseases and the general population as well (\url{https://tinyurl.com/trans-fats}, \url{https://tinyurl.com/trans-fat-cholestrol}).

Similarly, grilling a slice of meat can introduce carcinogenic agents \cite{cross2004meat} due to the animal fat dripping onto direct heat. However, grilling vegetables and fruits do not produce carcinogenic agents (\url{https://tinyurl.com/healthy-grill}). The process knowledge of cooking actions can aid the agent in learning general adverse effects due to specific combinations of cooking actions and ingredients. This process knowledge can aid the agent in learning to nudge any user towards healthy eating habits irrespective of the dietary guidelines for various chronic diseases. An agent learned by infusing the two kinds of process knowledge will be able to generate explanations, be interpretable, and thereby improve the safety aspect of meal advice.

\section*{Need for new Evaluation Metrics}
\noindent The precision of AI is not always a good indicator of clinical effectiveness. The area under the receiver operating characteristic curve (AUROC), another frequent metric, is not always the ideal indicator for clinical application. Such AI measures may be complex for physicians to comprehend or may not be clinically relevant. Furthermore, AI models have been assessed using a range of indices, including the F1 score, accuracy, and false-positive rate, which are indicators of distinct elements of AI's analytical ability. Understanding how complicated AI works necessitates a level of technical understanding not commonly seen among physicians.

AI models with process knowledge infusion require specialized metrics for evaluating the performance concerning safety and uncertainty, and risk handling. For instance, Stanford natural language inference, Multi-genre natural language inference, and others similar datasets can be used to create a learned evaluation metric to assess safety in generation by comparing the generated hypothesis with a premise (\url{https://tinyurl.com/NLI-datasets}) . In essence, safety and uncertainty and risk handling would require human evaluation, which is a mandate; these metrics are also equally important as they either involve: (a) annotators’ agreements/disagreements, (b) knowledge source, and (c) train deep language models on datasets that have data samples ordered by some relationships \cite{williams2017broad,camburu2018snli}.

\begin{itemize}
    \item [(a)] \textit{Average Number of Unsafe Matches: } This represents the average number of matches across all model-generated questions against a set consisting of utterances, lexical content, or ontology concepts used to describe harmful communication. Such a measure provides a range of means to impose safety checks that can be extracted from unstructured, semi-structured, and structured sources and domain experts. For example, named entities in the generated content could match against harmful concepts in a knowledge base or in a lexicon set containing harmful phrases (unigrams, bigrams, and trigrams). 
    \item [(b)] \textit{Perceived Risk Measure:} This is an annotator-in-the-loop metric to judge the model's stability in light of agreement and disagreement between the annotators, a notion of uncertainty and safety. It is composed of two components: (a) Penalty: A ratio of the count of misclassified samples over the count of those samples where the annotators disagree with each other. (b) Benefit:  A ratio of the count of samples where the model's predicted label agrees with some annotators (ignoring the disagreement between them) over the total number of annotators. Such a metric is efficient for controlling unsafe predictions as opposed to using statistical loss functions that quantify uncertainty in predictions and overwhelm the experts in the loop with re-annotations \cite{sawhney2022risk}. 
    \item [(c)]  \textit{Semantic Relations and Logical Agreement Measures:} These are trained metrics constructed using the RoBERTa model, a deep language model trained independently on sentence similarity and natural language inference GLUE tasks. These metrics have been introduced in a recent study by Gaur et al. that unites meta-information-guided passage retrievers and TEC for inducing logical ordering in the generations and preventing retrieval-augmented language models from hallucinations \cite{gaur2021iseeq}. Semantic Relation is a metric that counts the number of generations semantically similar to a user query over the total number of generations. The logical agreement score records the count when the current generated question entails a previously generated question. The score takes the sum of such counts and divides them by the number of generations.
\end{itemize}

\section*{Summary and Future Directions}
\noindent Real-world interactions between the users are not a single shot activity but rather a chain of exchanges involving procedural questions and responses; at a macroscopic level and the microscopic level, it comprises entities and actions that keep changing during a task-oriented conversation. This phenomenon can be well understood and controlled through a process of knowledge that represents a human’s mental model of conversation. In this article, through example use-cases in mental healthcare and food, we explained the notion of process knowledge that naturally concerns consistency, explainability, and interpretability in AI’s decision-making process. To the best of our knowledge, this article projects its role in pushing statistical AI to be safe, less uncertain, and risky in its classification and natural language generation tasks. With process knowledge, the AI model can support reasoning, which is essential to develop trust in stakeholders using the application in various downstream tasks. We showed various existing process knowledge and the methods that data-driven AI models can use. As a future direction, we envision the utility of process knowledge in personalization which is essential in developing interventional plans for patients with other mental health disorders (e.g., autism, aphasia) and developing food plans for patients with specific dietary needs. 
\section*{Acknowledgment}
This work was supported in part by National Science
Foundation (NSF) Award 2133842, “EAGER: Advancing
Neuro-symbolic AI with Deep Knowledge-infused
Learning.”







\bibliography{references.bib}
\bibliographystyle{IEEEtran}


\begin{itemize}
    \item Amit Sheth is the founding director of the AI Institute of South Carolina (AIISC: aiisc.ai),
NCR Chair and Professor of Computer Science and Engineering. Contact him at amit@sc.edu
    \item Manas Gaur recently completed Ph.D and is joining UMBC as assistant professor in the department of computer science. Contact him at mgaur@email.sc.edu
    \item Kaushik Roy, Revathy Venkataraman, and Vedant Khandelwal are current Ph.D. students advised by Prof. Sheth at AIISC. Contact them at \{kaushikr, revathy, vedant\}@email.sc.edu.
\end{itemize}









\end{document}